\documentclass[sigconf]{acmart}

\usepackage[false]{anonymous-acm}
\usepackage[inline,shortlabels]{enumitem}

\usepackage{multicol}

\usepackage{balance}

\usepackage[nameinlink]{cleveref}
\crefname{figure}{Fig.}{Figs.}
\Crefname{figure}{Figure}{Figures}
\crefname{equation}{Eq.}{Eqs.}
\Crefname{equation}{Equation}{Equations}
\crefname{section}{Section}{Sections}
\crefname{table}{Table}{Tables}
\crefname{appendix}{Appendix}{Appendices}
 
\newcommand{\eg}{e.g.,\ }

\newcommand{\ie}{i.e.,\ }

\newcommand{\Ereq}{E^\text{req}}  
\newcommand{\Erem}{E^\text{u}}  
\newcommand{\Eremt}{\Erem_t}  
\newcommand{\Eprovt}{E^\text{c}_t} 

\AtBeginDocument{%
  \providecommand\BibTeX{{%
    \normalfont B\kern-0.5em{\scshape i\kern-0.25em b}\kern-0.8em\TeX}}}

\copyrightyear{2026}
\acmYear{2026}
\setcopyright{cc}
\setcctype{by}
\acmConference[E-Energy '26]{The 17th ACM International Conference on Future and Sustainable Energy Systems}{June 22--25, 2026}{Banff, AB, Canada}
\acmBooktitle{The 17th ACM International Conference on Future and Sustainable Energy Systems (E-Energy '26), June 22--25, 2026, Banff, AB, Canada}
\acmDOI{10.1145/3744255.3811736}
\acmISBN{979-8-4007-2011-6/2026/06}

\graphicspath{{./images/}}

\begin{document}

\title[Forecasting what Matters: Decision‑Focused RL for Controlled EV Charging with Unknown Departure Times]{Forecasting what Matters: Decision‑Focused RL for\\
Controlled EV Charging with Unknown Departure Times}

\author{Giuseppe Gabriele}
\affiliation{%
  \institution{Ghent University -- imec}
  \city{Ghent}
  \country{Belgium}
}
\email{giuseppe.gabriele@ugent.be}
\orcid{0009-0008-4740-6370} 

\author{Fabio Pavirani}
\affiliation{
    \institution{Ghent University\,---\,imec}
    \city{Gent}
    \country{Belgium}}
\email{fabio.pavirani@ugent.be}
\orcid{0009-0005-7904-099X}

\author{Seyed~Soroush Karimi~Madahi}
\affiliation{%
  \institution{Ghent University -- imec}
  \city{Ghent}
  \country{Belgium}
}
\email{seyedsoroush.karimimadahi@ugent.be}
\orcid{XXXX-XXXX-XXXX-XXXX} 

\author{Chris Develder}
\affiliation{%
  \institution{Ghent University -- imec}
  \city{Ghent}
  \country{Belgium}
}
\email{chris.develder@ugent.be}
\orcid{0000-0003-2707-4176}

\renewcommand{\shortauthors}{G.\ Gabriele et al.}

\begin{abstract}
The recent growth of EV adoption poses challenges for power systems, including increased peak demand and potential grid instability.
Smart control of EV charging\,---\,\eg 
based on reinforcement learning (RL)\,---\,can alleviate these issues by learning temporal and contextual patterns from historical data.
Yet, in real-world scenarios, key features, such as departure time, often are unavailable.
This, in turn, makes it harder for an RL agent to learn and execute an effective charging policy.
To mitigate this uncertainty, a trained forecaster can approximate the unknown features from available data.
However, since these forecasting models are typically trained for accuracy (rather than their impact on a downstream agent's decision quality), their errors may propagate and hinder the overall performance of a controller that is using the forecasts. 
To 
avoid this, we propose a decision‑focused RL (DF-RL) framework in which the forecaster is trained \emph{end‑to‑end}, \ie with feedback from the charging policy actions taken by the RL agent.
Such joint training of both the forecaster and controller ultimately 
results in higher‑quality actions: 
our proposed DF-RL method yields superior charging decisions compared to other baselines, achieving up to a 14\% improvement in total reward and a 55\% reduction of unsupplied energy (i.e., charging that failed to happen because the EV already left), relative to the RL method without departure time forecasting.
\end{abstract}



\maketitle


\section{Introduction}
\label{sec:intro}
Electric vehicle (EV) charging scheduling is an increasingly important problem for the development of a cleaner and more resilient energy ecosystem, driven by the rapid growth in EV adoption~\cite{muratori2021rise}. 
By coordinating EVs\,---\,\ie deciding when and how vehicles are charged\,---\,energy demand can be balanced, costs can be reduced, and renewable power can be used more effectively. 
Optimizing EV charging thus is key in reducing overall energy expenses for both users and charging service providers.

Two prevalent approaches exist in the literature for EV charging scheduling.
The first is \emph{Model Predictive Control (MPC)}, 
which optimizes decisions over a short future horizon and updates them iteratively as new information becomes available.
In the EV context, it 
forecasts future demand, electricity prices, and vehicle availability, then computes the best charging plan while enforcing operational constraints \cite{8825773, 10089467}.
The second  widely used method 
is \emph{Reinforcement Learning (RL)}.
Thanks to its ability to learn optimal 
policies through interaction with the environment, 
 RL has demonstrated strong performance for a wide range of EV charging problems~\cite{madahi2025scalable, zhang2023distributed}.
Various objectives are considered, including load flattening~\cite{ 
, mosalli2025dynamic},
provision of grid services~\cite{park2022multi
}, and, as considered in our work, overall cost minimization~\cite{
, sharif2024smart}.
Since RL is data-driven and model-free, it can
overcome the challenges faced by MPC in dealing with uncertainties and nonlinear dynamics that are hard to model accurately.

A critical challenge for both RL and MPC is \emph{incomplete or unknown information}: 
e.g., the 
real departure time of the vehicle is typically unknown beforehand. 
This lack of information further complicates the problem, as 
the time flexibility available for EVs to shift their charging to off-peak hours is unknown.
Moreover, also user satisfaction and charging constraints are directly influenced by the departure time.
Only a limited number of works explicitly address this challenge: most studies assume access to 
exact departure time or 
perfect foresight state information.
In~\cite{mcclone2023hybrid}, a hybrid machine learning model is developed to forecast EV session durations, along with arrival times and energy demand. These forecasts are then fed into an MPC to minimize charging costs.
Also~\cite{shahriar2021prediction} proposed a machine learning-based framework to predict EV session duration and energy consumption.
In~\cite{wang2024semi} a semi-decentralized real-time charging scheduling scheme is put forward: a central operator uses a chance-constrained model to estimate aggregate charging energy and models uncertainties in EV users' behavior using a Gaussian mixture model.
Within RL‑based EV charging methods, augmented Lagrangian methods have been proposed to handle uncertainty and operational constraints by embedding constraint violations directly into the reward function~\cite{chen2022deepreinforcementlearningbasedcharging, chen2025deep}, rather than feeding the uncertain (predicted) departure time as part of the RL state.
Another relevant work is~\cite{tuchnitz2021development}, which incorporates 
a lookup table that stores historical EV information, from which departure times of each vehicle are inferred based on previously observed patterns. 
Yet, this approach is limited to direct interpolation over past observations, and 
this lack of flexibility
compromises generalization. 

When dealing with 
predicting unknown quantities for the sake of control,
\emph{decision-focused} (DF) learning offers an effective paradigm for integrating prediction and optimization. Unlike traditional approaches that prioritize predictive accuracy in isolation, DF learning 
accounts for the impact of predictions on the downstream decision-making problem.
DF-based solutions enhance the learning process by guiding predictions based on the quality of the resulting decisions rather than just prediction accuracy, allowing the control agent to 
further improve task performance~\cite{mandi2024decision, shah2022decision}.
%
Yet, only few studies have been conducted in EV charging control literature which use DF methods for predicting cost and energy charging demand~\cite{gu2024learningoptimizationpricebaseddemand, 10644254, 11090572}.
All these works formulate the EV charging problem within an optimization framework
and leverage decision-focused learning 
for predicting electricity price or energy demand. 
The main challenge in integrating DF learning with optimization-based methods is to calculate the gradient of task-aware loss functions with respect to the forecaster parameters, which can be highly non-trivial. To enable backpropagation through the downstream optimization problem, implicit differentiation can be applied to its Karush-Kuhn-Tucker (KKT) conditions (as in the papers cited above). Nevertheless, the inherent discontinuities in linear program outputs can pose difficulties for this approach.

Here, we address the challenge of dealing with a priori unknown EV departure times, while avoiding the complexity of and integrating DF learning with MPC and optimization-based methods.
We contribute a \textbf{decision-focused RL (DF-RL) framework} to effectively minimize EV charging costs under \textbf{uncertain EV departure times}.
We enrich the RL agent's state representation 
with a neural network-based forecast of the EV session duration.
In contrast to decision-focused MPC methods, our RL pipeline allows the forecaster 
to be 
trained directly from optimizing the control objective: the forecaster 
and (RL) control agent are thus updated in a  \emph{decision-focused} way and trained \emph{end-to-end}.
To the best of our knowledge, the application of the decision‑focused learning in RL
is unique in the context of EV charging scheduling. 

\section{Problem Formulation}
\label{sec:probform}
We formalize the EV charging control problem as a Markov Decision Process (MDP)~\cite{puterman1990markov}: $\mathcal{M} = (\mathcal{S},\, \mathcal{A},\, \mathcal{P},\, \rho)$, where $\mathcal{S}$ defines the observable state space,  $\mathcal{A}$ the action space, $\mathcal{P}$ the state transition function (mapping current state and action to the next state), and $\rho$ the reward function. 
We consider 
sequential EV charging sessions, each
one defined by arrival time, departure time, and energy demand (to fully charge the battery; in kWh). 
We adopt a discrete-time setting, with a fixed time resolution $\Delta t$ (30\,min in our study).

The goal is to find a policy 
$\pi: \mathcal{S} \rightarrow \mathcal{A}$
that gives the action to perform, based on a given state as input, to maximize the expected cumulative reward.
For such a given policy $\pi$, we consider the Q-function $Q^\pi$ that quantifies the long‑term value of taking action $a_t$ when in state $s_t$ (with immediate reward $\rho(s_t,a_t)$) and continuing to follow 
the policy $\pi$ (i.e., $a_t' = \pi(s_t')$ from $t'=t+1$ onwards)~\cite{busoniu2017reinforcement}:
\begin{equation}
Q^{\pi}(s_t,a_t)
=
\mathbb{E}_{\pi}
\left[
\rho(s_t,a_t)
+
\sum_{k=1}^{K} \gamma^{k} \, \rho(s_{t+k},a_{t+k}) \right]
\end{equation}

%
Specifically, the agent’s \textbf{state} 
comprises two parts: 
\begin{equation}
    s_t = (m_t, \hat d_t) \in \mathcal{S}.  \label{eq:enhanced_state}
\end{equation}
The first component $m_t$ comprises known and measurable variables:
$m_t = (p_t, \Eremt, h_t)$, where $p_t$ is the 
electricity price (€/kWh) for that timestep,
and $\Eremt$ is the energy (kWh) that the EV still needs
to be fully charged.
Note that we assume to know the total energy that the EV needs to be charged with upfront, and denote it as $\Ereq$ (thus $\Eremt = \Ereq$ at the time or arrival of the EV).
Finally, 
$h_t \in \mathcal{H} \triangleq \{0, \Delta t, 2\Delta t, ..., 24-\Delta t\}$ represents the time of the day.

The second component $\hat d_t$ represents the remaining charging session duration, which in practice we do not know exactly.
To estimate it, we forecast the total session duration $\hat {T}_t$ at each time step (\cref{eq:forecaster_time}). Forecasting at every timestep allows $\hat {T}_t$ to take into account the most recent information about the unmet energy.
The remaining session duration is then obtained 
using \cref{eq:session_duration}.
\begin{align}
    \hat {T}_t &= f_{\psi}(x_t)    \label{eq:forecaster_time}
    \\
    \hat d_t &= \hat T_t - T^{\text{elapsed}}_t       \label{eq:session_duration}
\end{align}
Our forecaster $f_{\psi}(.)$ takes the following inputs:
\begin{equation}
    x_t = (z_t, \Eprovt, \Eremt),
    \label{eq:input_fc}
\end{equation}
where %
$z_t$ is relevant time information (\eg including the hour of the day, day of the week, season),\footnote{In our experiments, based on data analysis of the real-world EV charging sessions, we set $z_t = (h_t, w_t)$ where $h_t$ is the time of the day, and $w_t \in [0,1]$ a binary variable indicating whether it is weekend or not.}
and $\Eprovt$ is the energy the EV has been charged with so far at time $t$.

We consider a binary \textbf{action} space: the charging action $a_t \in \mathcal{A}$ is to either charge at full power
($a_t=P_c$, where $P_c$ is a priori fixed charging power) or 
to remain idle ($a_t=0$). 

In our problem, the state \textbf{transition function} includes some stochasticity, mainly through the time-varying price $p_t$ and the session duration.
Part of the transition function is defined by \cref{eq:energy_update}, which shows the update of the remaining energy for the session
\begin{equation}
    E^u_{t+1} = \Eremt - E_t,
    \label{eq:energy_update}
\end{equation}
where $E_t$ indicates the energy delivered within the slot, defined as:
\begin{equation*}
    E_t =  \text{min}(a_t \cdot \Delta t, \Eremt).
\end{equation*}


We define the \textbf{reward} $r_t \triangleq \rho(s_t, a_t)$ 
for training and evaluation:
\begin{equation}
r_t = -\,p_t \cdot E_t \;-\; C_t \cdot \mathbb{I}\left\{\text{EV left at timestep } t \right\}, 
\label{eq:rl_reward}
\end{equation}
where $\mathbb{I}\{.\}$ is the indicator function, and $C_t$ an extra penalty cost, 
which through the indicator function only applies upon departure: 
\begin{equation}
C_t 
= 
\begin{cases}
\left(\lambda \cdot \Eremt \right)^2 + C_{\mathrm{fix}}, 
& \text{if } \Eremt > 0 \text{ (EV not fully charged)}, \\
0, 
& \text{otherwise}.
\end{cases} 
\end{equation}
The first term of \cref{eq:rl_reward} captures the charging cost incurred at time step $t$.
The second term, 
adds a penalty cost when a vehicle is not fully charged when it leaves. 
That penalty comprises a fixed cost $C_\text{fix}$ and a squared penalty for the unmet charging demand upon departure (\ie $\Eremt$), regularized by a constant $\lambda$. 

\section{Methodology}
\label{sec:methodology}

In the proposed EV charging scheduling framework, as defined formally by the MDP stated above, 
we use Soft Actor Critic (SAC)~\cite{haarnoja2018softactorcriticoffpolicymaximum} as the RL algorithm. 
SAC is an off‑policy actor-critic algorithm~\cite{grondman2012survey} 
that encourages exploration by maximizing not only expected return, but also the entropy of the policy.
That trade-off between reward maximization and policy entropy, is controlled by a temperature parameter $\alpha$, which is learned online via automatic entropy tuning based on the current policy entropy.
In our \textbf{proposed DF-RL method}, RL incorporates a regression forecaster model $f_{\psi}$, which augments the state representation by providing additional predictive information (in our case the EV's session duration).
%
Following the DF concept, the forecaster is trained jointly with the RL objective, to not only focus on accurate forecasts, but also optimizing the RL policy's obtained reward.
Specifically, the forecaster is updated during training by minimizing the following loss function:
\begin{equation}
 \mathcal{L}_{\text{total}}
    = \beta\,\mathcal{L}_{R} \;+\; (1-\beta)\,\mathcal{L}_{DF},
    \qquad \beta \in [0,1].
    \label{eq:betalosses}
\end{equation}
Here, the first term, $\mathcal{L}_{R}$, updates the forecaster 
based on a regression loss function $\ell(\cdot)$ for the deviation between the forecasted session duration $f(x_{t})$ and the real session duration $T_\text{tot}$ of the EV.
\footnote{In our implementation, we use Mean Squared Error (MSE) loss.}

The second term, $\mathcal{L}_{\mathrm{DF}}$, corresponds to the decision‑focused component and leverages the actor’s decision to guide the update of the forecaster:
\begin{equation}
\mathcal{L}_{\text{DF}}
= \mathbb{E}_{s}\left[\sum_{a \in \mathcal{A}} \pi(a\mid s)\,\big(\alpha\,\log \pi(a\mid s) - A(s,a)\big)\right].
\end{equation}
In this formulation, $A(s,a)$ denotes the advantage function obtained from the standardized Q-value
~\cite{6392457}, 
\ie the action‑normalized Q‑value, and its policy‑weighted expectation.\footnote{Note that here we assume the policy function $\pi(s,a)$ to return the probability to take action $a$ when in given state $s$.}

Within this framework, the parameters of the DF-RL algorithm are updated at the end of each charging session, including the forecaster parameters, optimized via the DF loss defined in \cref{eq:betalosses}. 

Note that through the weight parameter $\beta$ in \cref{eq:betalosses}, we can modulate the degree of 
updating the forecaster based on the DF objective:
setting $\beta=0$ results in a purely DF-based update, whereas $\beta=1$ corresponds to a standard regression‑based update with respect to the target value.

Now that we 
explained 
our DF-RL framework, in the next section, we investigate the benefits of a DF forecaster (and thus the resulting DF-RL policy) over 
the forecaster trained using a regression loss and integrated with RL.

\section{Experiments}
\label{sec:exper}
\begin{table*}
  \caption{
  Comparison of our proposed decision-focused RL (DF-RL) approach vs.\ conventional RL and business-as-usual (BAU) baselines. Metrics are averages $\pm$ standard deviation over 4 experiment runs, on a test set of $N=\text{120}$ EV charging sessions.}
  \small
  \label{tab:results}
    \begin{tabular}{llcccccc}
    \toprule
    && \multicolumn{3}{c}{User-focused metrics} & \multicolumn{2}{c}{RL algorithm metrics} \\
    \cmidrule(lr){3-5}\cmidrule(lr){6-7}
    \multicolumn{2}{l}{Algorithm} & Charging Cost (€) & \# Unsatisfied EVs & Unmet Demand (\%) & Total Reward & 
    Penalty Cost \\
    \midrule
    \midrule
    \multicolumn{2}{l}{BAU (immediate charging)} 
    & 168.43 & 0.0 & 0.0 & $-$168.43 & 0.0 \\
    \midrule
    
    \multicolumn{2}{l}{RL (no forecast)}  
    & \phantom{0}60.94\,$\pm$\,14.06 
        & 33\,$\pm$\,\phantom{0}4.95 & 14.7\,$\pm$\,9.5 & $-$166.88\,$\pm$\,22.20 & 105.94\,$\pm$\,36.26 \\
    
    \multicolumn{2}{l}{RL w/ Forecaster ($\beta=1$)} 
    & \phantom{0}93.42
\,$\pm$\,14.06 & 33\,$\pm$\,11.67 & \phantom{0}7.3\,$\pm$\,5.2 & $-$151.00\,$\pm$\,22.20 & \phantom{0}57.58\,$\pm$\,36.26 \\
    
    \multicolumn{2}{l}{RL w/ Real Departure}
    & 102.17\,$\pm$\,\phantom{0}3.66 & \phantom{0}5\,$\pm$\,\phantom{0}4.71 & \phantom{0}0.6\,$\pm$\,0.6 
        & $-$115.48\,$\pm$\,11.09 & \phantom{0}13.31\,$\pm$\,\phantom{0}8.91 \\
    
    \midrule
    DF-RL & $\beta = 0$ 
    & \phantom{0}55.70\,$\pm$\,13.90 & 48\,$\pm$\,\phantom{0}1.50 & 24.9\,$\pm$\,1.3 & $-$177.46\,$\pm$\,\phantom{0}9.01 & 121.76\,$\pm$\,\phantom{0}4.89 \\
    
    DF-RL & $\beta= 0.2$ 
    & \phantom{0}93.79\,$\pm$\,12.36 & 21\,$\pm$\,\phantom{0}8.14 & \phantom{0}6.3\,$\pm$\,4.3 & $-$142.85\,$\pm$\,\phantom{0}6.59 & \phantom{0}51.16\,$\pm$\,16.87 \\
    
    DF-RL & $\beta= 0.4$ 
    & \phantom{0}91.90\,$\pm$\,14.46 & 24\,$\pm$\,13.32 & \phantom{0}6.6\,$\pm$\,5.6 & $-$142.85\,$\pm$\,14.02 & \phantom{0}50.96\,$\pm$\,26.04 \\
    
    DF-RL & $\beta= 0.6$ 
    & \phantom{0}94.69\,$\pm$\,15.45 & 24\,$\pm$\,12.42 & \phantom{0}6.6\,$\pm$\,5.6 & $-$142.90\,$\pm$\,12.61 & \phantom{0}48.21\,$\pm$\,27.73 \\
    
    DF-RL & $\beta= 0.8$ 
    & \phantom{0}92.62\,$\pm$\,14.36 & 27\,$\pm$\,10.26 & \phantom{0}7.3\,$\pm$\,4.0 & $-$144.14\,$\pm$\,\phantom{0}6.82 & \phantom{0}51.52\,$\pm$\,20.99 \\
    \bottomrule
  \end{tabular}
\end{table*}


Our experiments are based on a real-world dataset of EV charging sessions, from which we extract vehicle arrival times, session durations, and charging requirements.
The sessions are collected from an EV parking station located at a hospital\textanon{\footnote{This was collected at a public parking facility at a hospital, in the frame of the European H2020 project RENergetic~\url{https://cordis.europa.eu/project/id/957845}}.}.
From this dataset, the 20 most frequent EV users are considered
to model EV features 
using statistical distributions to generate realistic simulation samples.
We thus construct a training dataset of 350 sessions, and will test on $N=\text{120}$ sessions to report results.
For all sessions, we assume a fixed charging power of $P_c = \text{6.5}$\,kW.
Since we focus on time departure uncertainty, we 
assume a single deterministic time-varying price profile for all sessions, shown in Appendix~\ref{App1}.
%

To assess the performance of DF-RL, we compare several variants, ranging $\beta$ in [0,1]. 
We further pretrain the forecaster purely based on minimizing regression loss, separate from the training data used for the RL agent.
To assess the relevance of the RL agent knowing the session duration (or not), we further provide a baseline where it does not have any information (`no forecast'; \ie $\hat d_t$ is omitted from the state) or has perfect information (`real departure').
To gauge the performance of these RL-based solutions, we also provide a business-as-usual (BAU) baseline where the EV starts charging 
immediately upon arrival.
Finally, we note that the RL agents are trained for 200 episodes (an episode being one pass over the entire training set), the discount factor is set to $\gamma=\text{0.99}$, and cost parameters are $C_\text{fix}=\text{2.1}$, 
$\lambda=\text{0.041}$.


To evaluate the performance of these DF-RL approaches and 
mentioned
baselines, we report metrics relevant for the EV users:
\begin{enumerate*}[(i)]
\item the total charging cost of all EVs (`charging cost'),
\item number of sessions with incomplete charging (`unsatisfied EVs'), and
\item the fraction of the unmet charging load (`unmet demand').
\end{enumerate*}
The latter \emph{unmet demand} is more specifically calculated as the 
percentage
of kWh that were requested during the EV session ($\Ereq$), but were eventually not supplied by the time it departs ($E^\text{unmet}$), averaged over all $N$ sessions:
\begin{equation}
\text{Unmet demand}
= \frac{1}{N}\sum_{i=1}^{N} 
\frac{E^{\text{unmet}}_{i}}{E^{\text{req}}_{i}}.
\label{eq:pved}
\end{equation}
To provide insight in the RL algorithm's operation, we further list their reward and penalty costs, \ie
\begin{enumerate*}[(i)]
    \item the cumulative reward defined in \cref{eq:rl_reward}, over all test charging sessions, and
    \item the total penalty cost, $C_t$, again summed over all sessions.
\end{enumerate*}

The resulting metrics are reported in \cref{tab:results}. 
We notice that all vanilla RL-based methods outperform the BAU solution in terms of total reward, primarily due to the high charging costs incurred by the BAU strategy.
Focusing on the vanilla RL methods, we note the expected benefit of having departure time information: RL with the conventional forecaster ($\beta = 1$) almost halves the unmet demand,
compared to the RL without any forecaster.
Still, the forecast errors imply that there is still a meaningful performance gap compared to having perfect information (`RL w/ Real Departure'), where unmet demand is virtually eliminated.\footnote{
Even then it is not 0, since we use
a penalty factor for constraint satisfaction, applying a soft penalty rather than enforcing a hard constraint on the agent.}

Looking at our decision-based forecast methods (DF-RL), while they still use the same inputs as conventional forecasting, we note that DF-RL 
manages to further reduce unmet demand, while also achieving a lower charging cost (see $\beta = 0.2$ or $0.4$):
out of the 120 EVs in the test set, an additional $\sim$10 are fully charged compared to RL using conventional forecasts.
We note that lower $\beta$ (e.g., 0.2) results in slightly higher costs\,---\,similar to $\beta = 1$\,---\,but fewer unmet charging demands, while higher $\beta$ (e.g., 0.8) reduces cost at the expense of more unsatisfied EVs. This reflects the trade-off: lower $\beta$ prioritizes meeting energy demand, increasing electricity use. Additionally, higher $\beta$ leads to greater cost variability, whereas $\beta = 0.2$ produces more stable outcomes.
In terms of the RL algorithm's internal metrics, using DF-RL 
increases total reward by 
15\%, 13\%, and 5\%, 
compared to, respectively, BAU, RL without departure time forecasts, and RL with conventional (non-DF) forecasting. 
Overall, all DF-RL variants with $0<\beta<1$ obtain promising results.\footnote{For $\beta=0$, which amounts to neglecting forecasting accuracy (see \cref{eq:betalosses}), DF-RL unexpectedly performs poorly:
without forecaster loss guidance, the forecaster and RL agent still converge but requires more exploration and additional training iterations.}

\begin{figure}
    \centering
    \includegraphics[width=0.9\columnwidth]{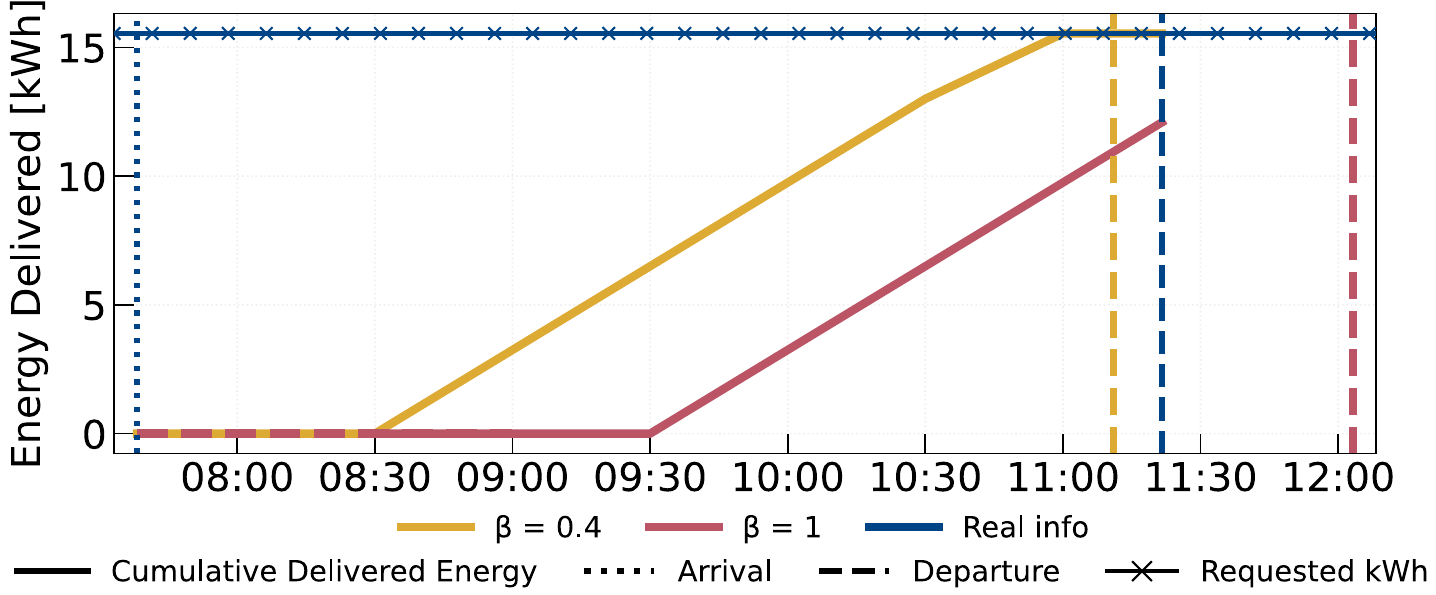}
        \caption{Exemplary charging session for DF-RL (with $\beta=\text{0.4}$) compared to traditional RL (non-DF forecaster, with $\beta = \text{1}$).}
    \label{fig:charging_example}
\end{figure}

As already noted in \cref{tab:results}, the main benefit of our proposed DF-RL approaches is that it reduces the number of EVs that fail to complete charging (`unsatisfied EVs').
This suggests that our DF forecasts do generate signals that are more informative for the RL agent’s policy learning (while not necessarily producing more accurate predictions).
A particular example of this behavior is illustrated in \cref{fig:charging_example}, which compares the same EV charging session under DF‑RL with $\beta=0.4$ and RL with 
the traditional forecaster ($\beta = 1$).
We note that the decision‑focused forecaster predicts a slightly shorter session duration (ending around 11:10) than the true one (around 11:20).
This conservative prediction effectively encourages the RL agent to start charging early enough (at 8:30), allowing the EV to complete charging before its actual departure.
In contrast, 
the traditional 
forecaster overestimates the available session duration (predicting its end around 12:05), leading to delayed charging and an incomplete charging session.
Overall, these 
results 
thus demonstrate our proposed DF-RL approach's effectiveness. 

\section{Conclusions and Future Work}
We introduced a decision focused reinforcement learning (DF-RL) framework, to solve the problem of dealing with unknown information (focusing particularly on departure times), in coordinated EV charging scenarios. 
Our DF-RL solution introduces a forecaster that augments the RL agent's state and is jointly trained using feedback from that agent.
In the considered EV charging control problem, the forecaster predicts the anticipated EV's departure time (learned from historical behavior).
In our DF-RL approach, that prediction is optimized to 
improve charging decisions, rather than solely focusing on prediction accuracy (based on MSE minimization).
We benchmarked our DF-RL approach against RL with such conventional forecasting (as well as boundary baselines comprising 
a business-as-usual ``charge‑when‑plugged'' scenario,
and an RL agent without any departure time information). 
This comparison demonstrated the benefit of our DF-RL in terms of a 5\% improvement on the total reward, and 14\% less of the total unmet energy, compared to the model with the conventional forecaster.

Future work will focus on extending the proposed DF-RL framework to a more realistic scenario with different daily price profiles, incorporating more expressive forecasting models capable of capturing temporal patterns such as time-of-day effects and seasonality. In addition, we plan to investigate cooperative extensions of the proposed framework, where multiple vehicles 
are controlled 
to better handle grid constraints and system-level limitations, ultimately enabling a scalable and more efficient charging solution.




\begin{acks}
\small
This research was partly funded by the Flemish Government
through the ``Onderzoeksprogramma Artifici\"{e}le Intelligentie (AI) Vlaanderen'' programme, and by the EU through the Horizon Europe project BlueBird (\url{https://bluebirdproject.eu/} -- grant agreement no.\ 101192452)
\end{acks}
\balance 
\bibliographystyle{ACM-Reference-Format}
\bibliography{references.bib}

@article{muratori2021rise,
  title={The rise of electric vehicles—2020 status and future expectations},
  author={Muratori, Matteo and Alexander, Marcus and Arent, Doug and Bazilian, Morgan and Cazzola, Pierpaolo and Dede, Ercan M and Farrell, John and Gearhart, Chris and Greene, David and Jenn, Alan and others},
  journal={Progress in Energy},
  volume={3},
  number={2},
  pages={022002},
  year={2021},
  publisher={IOP Publishing}
}

@inproceedings{
madahi2025scalable,
title={Scalable Attention-based Reinforcement Learning Method for Multi-asset Control},
author={Seyed Soroush Karimi Madahi and Giuseppe Gabriele and Bert Claessens and Chris Develder},
booktitle={ICML 2025 CO-BUILD  Workshop on Computational Optimization of Buildings},
year={2025},
url={https://openreview.net/forum?id=3h0v1Ht73L}
}

@inproceedings{mosalli2025dynamic,
  title={Dynamic Load Balancing for EV Charging Stations Using Reinforcement Learning and Demand Prediction},
  author={Mosalli, Hesam and Sanami, Saba and Yang, Yu and Yeh, Hen-Geul and Aghdam, Amir G},
  booktitle={2025 IEEE International systems Conference (SysCon)},
  pages={1--7},
  year={2025},
  organization={IEEE}
}

@article{shah2022decision,
  title={Decision-focused learning without decision-making: Learning locally optimized decision losses},
  author={Shah, Sanket and Wang, Kai and Wilder, Bryan and Perrault, Andrew and Tambe, Milind},
  journal={Advances in Neural Information Processing Systems},
  volume={35},
  pages={1320--1332},
  year={2022}
}

@article{park2022multi,
  title={Multi-agent deep reinforcement learning approach for EV charging scheduling in a smart grid},
  author={Park, Keonwoo and Moon, Ilkyeong},
  journal={Applied energy},
  volume={328},
  pages={120111},
  year={2022},
  publisher={Elsevier}
}

@article{tuchnitz2021development,
  title={Development and evaluation of a smart charging strategy for an electric vehicle fleet based on reinforcement learning},
  author={Tuchnitz, Felix and Ebell, Niklas and Schlund, Jonas and Pruckner, Marco},
  journal={Applied Energy},
  volume={285},
  pages={116382},
  year={2021},
  publisher={Elsevier}
}

@article{sharif2024smart,
  title={Smart EV charging with context-awareness: Enhancing resource utilization via deep reinforcement learning},
  author={Sharif, Muddsair and Seker, Huseyin},
  journal={IEEE Access},
  volume={12},
  pages={7009--7027},
  year={2024},
  publisher={IEEE}
}

@article{zhang2023distributed,
  title={Distributed training and distributed execution-based Stackelberg multi-agent reinforcement learning for EV charging scheduling},
  author={Zhang, Jin and Che, Liang and Shahidehpour, Mohammad},
  journal={IEEE Transactions on Smart Grid},
  volume={14},
  number={6},
  pages={4976--4979},
  year={2023},
  publisher={IEEE}
}

@misc{chen2022deepreinforcementlearningbasedcharging,
      title={A Deep Reinforcement Learning-Based Charging Scheduling Approach with Augmented Lagrangian for Electric Vehicle}, 
      author={Guibin. Chen and Xiaoying. Shi},
      year={2022},
      eprint={2209.09772},
      archivePrefix={arXiv},
      primaryClass={cs.AI},
      url={https://arxiv.org/abs/2209.09772}, 
}

@article{chen2025deep,
  title={A deep reinforcement learning-based charging scheduling approach with augmented Lagrangian for electric vehicles},
  author={Chen, Guibin and Yang, Lun and Cao, Xiaoyu},
  journal={Applied Energy},
  volume={378},
  pages={124706},
  year={2025},
  publisher={Elsevier}
}

@article{mandi2024decision,
  title={Decision-focused learning: Foundations, state of the art, benchmark and future opportunities},
  author={Mandi, Jayanta and Kotary, James and Berden, Senne and Mulamba, Maxime and Bucarey, Victor and Guns, Tias and Fioretto, Ferdinando},
  journal={Journal of Artificial Intelligence Research},
  volume={80},
  pages={1623--1701},
  year={2024}
}

@misc{gu2024learningoptimizationpricebaseddemand,
      title={Learning and Optimization for Price-based Demand Response of Electric Vehicle Charging}, 
      author={Chengyang Gu and Yuxin Pan and Ruohong Liu and Yize Chen},
      year={2024},
      eprint={2404.10311},
      archivePrefix={arXiv},
      primaryClass={eess.SY},
      url={https://arxiv.org/abs/2404.10311}, 
}

@misc{haarnoja2018softactorcriticoffpolicymaximum,
      title={Soft Actor-Critic: Off-Policy Maximum Entropy Deep Reinforcement Learning with a Stochastic Actor}, 
      author={Tuomas Haarnoja and Aurick Zhou and Pieter Abbeel and Sergey Levine},
      year={2018},
      eprint={1801.01290},
      archivePrefix={arXiv},
      primaryClass={cs.LG},
      url={https://arxiv.org/abs/1801.01290}, 
}

@INPROCEEDINGS{10644254,
  author={Gu, Chengyang and Pan, Yuxin and Liu, Ruohong and Chen, Yize},
  booktitle={2024 American Control Conference (ACC)}, 
  title={Learning and Optimization for Price-Based Demand Response of Electric Vehicle Charging}, 
  year={2024},
  volume={},
  number={},
  pages={3625-3630},
  doi={10.23919/ACC60939.2024.10644254}}

@INPROCEEDINGS{11090572,
  author={Hao, Wenxian and Wang, Jingxiang and Wang, Zhaojian},
  booktitle={2025 37th Chinese Control and Decision Conference (CCDC)}, 
  title={Day-Ahead V2G Station Arbitrage Scheduling: A Decision-Focused Approach}, 
  year={2025},
  volume={},
  number={},
  pages={1531-1537},
  doi={10.1109/CCDC65474.2025.11090572}}

@article{puterman1990markov,
  title={Markov decision processes},
  author={Puterman, Martin L},
  journal={Handbooks in operations research and management science},
  volume={2},
  pages={331--434},
  year={1990},
  publisher={Elsevier}
}

@book{busoniu2017reinforcement,
  title={Reinforcement learning and dynamic programming using function approximators},
  author={Busoniu, Lucian and Babuska, Robert and De Schutter, Bart and Ernst, Damien},
  year={2017},
  publisher={CRC press}
}

@ARTICLE{8825773,
  author={Al-Ogaili, Ali Saadon and Tengku Hashim, Tengku Juhana and Rahmat, Nur Azzammudin and Ramasamy, Agileswari K. and Marsadek, Marayati Binti and Faisal, Mohammad and Hannan, Mahammad A.},
  journal={IEEE Access}, 
  title={Review on Scheduling, Clustering, and Forecasting Strategies for Controlling Electric Vehicle Charging: Challenges and Recommendations}, 
  year={2019},
  volume={7},
  number={},
  pages={128353-128371},
  doi={10.1109/ACCESS.2019.2939595}}

@ARTICLE{10089467,
  author={Yang, Lei and Geng, Xinbo and Guan, Xiaohong and Tong, Lang},
  journal={IEEE Transactions on Automation Science and Engineering}, 
  title={EV Charging Scheduling Under Demand Charge: A Block Model Predictive Control Approach}, 
  year={2024},
  volume={21},
  number={2},
  pages={2125-2138},
  doi={10.1109/TASE.2023.3260804}}

@article{grondman2012survey,
  title={A survey of actor-critic reinforcement learning: Standard and natural policy gradients},
  author={Grondman, Ivo and Busoniu, Lucian and Lopes, Gabriel AD and Babuska, Robert},
  journal={IEEE Transactions on Systems, Man, and Cybernetics, part C (applications and reviews)},
  volume={42},
  number={6},
  pages={1291--1307},
  year={2012},
  publisher={IEEE}
}

@ARTICLE{6392457,
  author={Grondman, Ivo and Busoniu, Lucian and Lopes, Gabriel A. D. and Babuska, Robert},
  journal={IEEE Transactions on Systems, Man, and Cybernetics, Part C (Applications and Reviews)}, 
  title={A Survey of Actor-Critic Reinforcement Learning: Standard and Natural Policy Gradients}, 
  year={2012},
  volume={42},
  number={6},
  pages={1291-1307},
  doi={10.1109/TSMCC.2012.2218595}}

@article{mcclone2023hybrid,
  title={Hybrid machine learning forecasting for online mpc of work place electric vehicle charging},
  author={McClone, Graham and Ghosh, Avik and Khurram, Adil and Washom, Byron and Kleissl, Jan},
  journal={IEEE Transactions on Smart Grid},
  volume={15},
  number={2},
  pages={1891--1901},
  year={2023},
  publisher={IEEE}
}

@article{shahriar2021prediction,
  title={Prediction of EV charging behavior using machine learning},
  author={Shahriar, Sakib and Al-Ali, Abdul-Rahman and Osman, Ahmed H and Dhou, Salam and Nijim, Mais},
  journal={Ieee Access},
  volume={9},
  pages={111576--111586},
  year={2021},
  publisher={IEEE}
}

@article{wang2024semi,
  title={A semi-decentralized real-time charging scheduling scheme for large EV parking lots considering uncertain EV arrival and departure},
  author={Wang, Weilun and Wu, Lei},
  journal={IEEE Transactions on Smart Grid},
  volume={15},
  number={6},
  pages={5871--5884},
  year={2024},
  publisher={IEEE}
}


\appendix

\section{Price Profile} \label{App1}

\begin{figure}[h]
    \centering
    \includegraphics[width=\columnwidth]{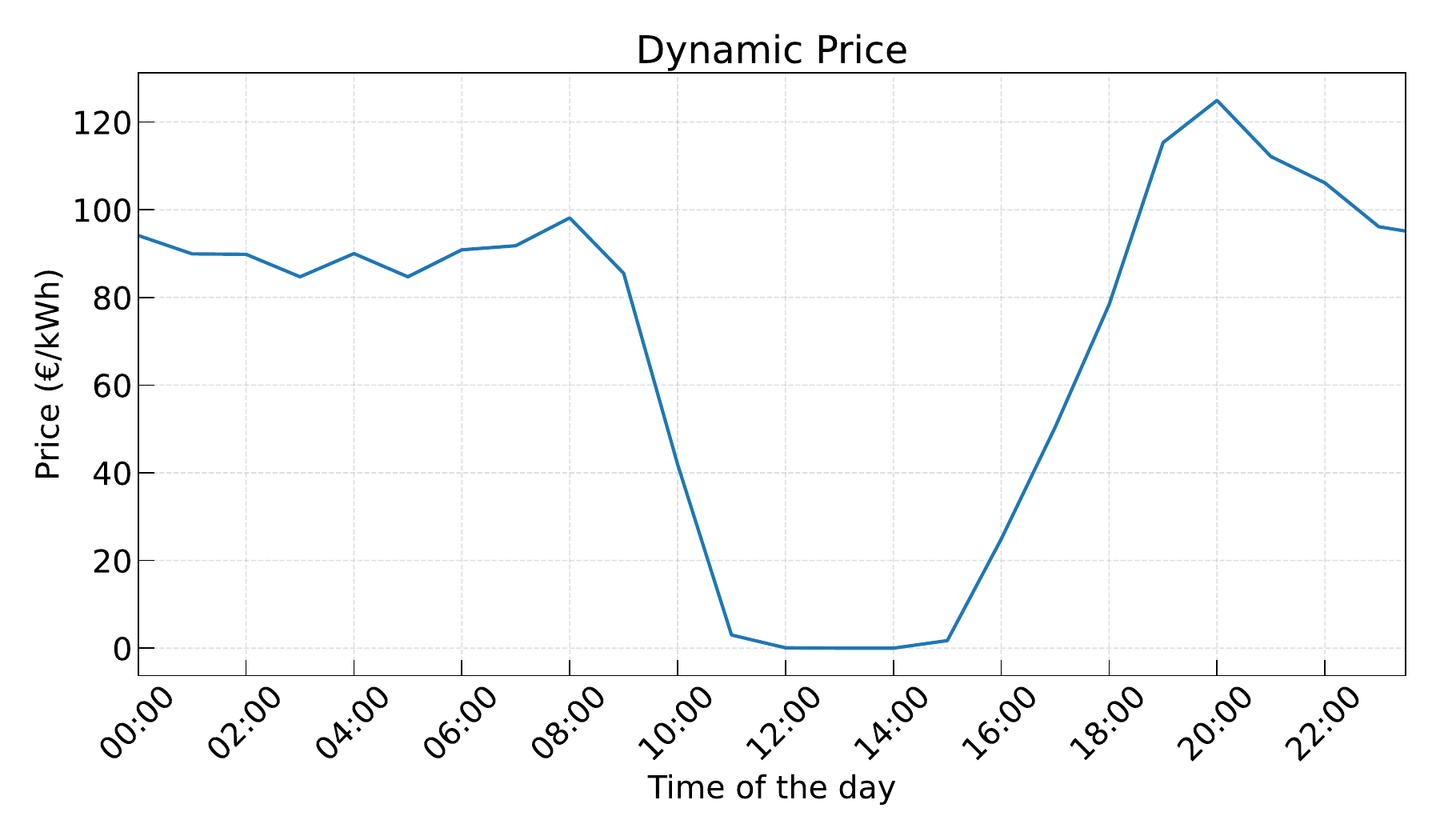}
        \caption{Daily price profile considered in our experiments.}
    \label{fig:price-profile}
\end{figure}

\end{document}